# Uncertainty Aware Active Learning for Reconfiguration of Pre-trained Deep Object-Detection Networks for New Target Domains

**Jiaming Na, Varuna De-Silva**
Loughborough University London
{j.na2, v.d.de-silva}@lboro.ac.uk

## Abstract

Object detection is one of the most important and fundamental aspects of computer vision tasks, which has been broadly utilized in pose estimation, object tracking and instance segmentation models. To obtain training data for object detection model efficiently, many datasets opt to obtain their unannotated data in video format and the annotator needs to draw a bounding box around each object in the images. Annotating every frame from a video is costly and inefficient since many frames contain very similar information for the model to learn from. How to select the most informative frames from a video to annotate has become a highly practical task to solve but attracted little attention in research. In this paper, we proposed a novel active learning algorithm for object detection models to tackle this problem. In the proposed active learning algorithm, both classification and localization informativeness of unlabelled data are measured and aggregated. Utilizing the temporal information from video frames, two novel localization informativeness measurements are proposed. Furthermore, a weight curve is proposed to avoid querying adjacent frames. Proposed active learning algorithm with multiple configurations was evaluated on the MuPoTS dataset [1] and FootballPD dataset.

## 1 Introduction

Object detection models have been widely used in monitoring complex traffic scenes [2], [3]; autonomous driving [4], [5]; human face recognition [6], [7] and sports data analytics [8], [9]. The object detection task is to detect all the instances of interested classes from the input image. Since various computer vision tasks are built upon an object detection model, improving the object detection accuracy has always been a crucial computer vision objective.

Most of the object detection models [10], [11] consist of two components: localization and classification. During the localization phase, a set of bounding box candidates are proposed based on the likelihood of them containing an object. Each of the bounding box candidates are classified with a class score in the classification phase. Lastly, the duplicated bounding boxes are removed and only the most likely ones are left as the final detection.

The state-of-the-art object detection models are trained with large datasets such as PASCAL Visual Object Classes dataset [12] and Common Objects in Context dataset [13]. The training process usually takes multiple days on GPUs depending on the model architecture [10], [11]. Some of pre-trained weights of these neural networks are available to public. However, directly using the pre-trained weights on new input data might lead to two practical issues: 1) the input data contains undetectable classes that was not included in the training data and 2) the accuracy is much lower compared to the reported evaluation accuracy, since the feature distribution of the new input data is drastically different to the training dataset [14]. In such situation, user usually adept the architecture of these object detection models and train it with their own dataset to reach certain accuracy.

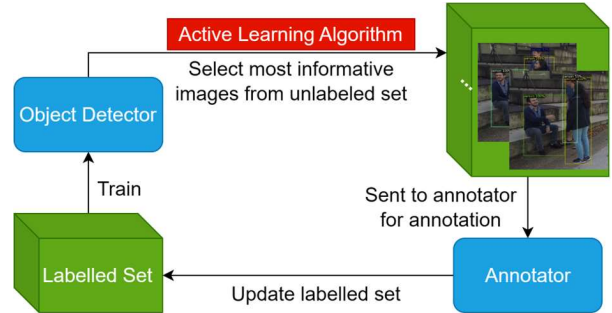

Figure 1. One active learning iteration for object detection model.

To train a deep learning model for object detection requires massive amount of training data. The key advantage of deep learning is the capability of extracting features from high-dimensional data. However, high-dimensional data is always more costly to annotate [15]. In the circumstance that there are unlabelled data available and the annotation task for each example is easy to perform, active learning is a labour-efficient solution to train a deep learning model more effectively within a supervised setting [15]. As shown in Figure 1, active learning algorithms let the deep learning model query specific data examples based on query strategy and the human annotator only labels the selected examples for future training. By combining active learning and deep learning, it is possible to achieve a higher performance with lower data annotation cost compared to passive learning.

One common way for unannotated image data collection is through video since shooting a video is less time consuming compared to manually collect images. The occlusion and motion blur induced by the nature of videos can make the object detection task more difficult. However, by properly annotate these data examples, the final trained model can also be more prone to noises [16].

In this paper, we propose an active learning algorithm to efficiently query frames from an unannotated video data for annotation, so that the object detection model could reach a desired accuracy with the least number of frames annotated for training.

An object detection model usually contains two components: localization and classification. Multiple query strategies for measuring data informativeness for the classification component in the object detection model has been proposed [17], [18]. However, [19], [20] has shown that the data informativeness for the localization component is also crucial to the performance of an active learning algorithm. Such informativeness is a difficult to measure since object detection models performs the localization task differently. Most previous active learning algorithms measured the data localization informativeness by modifying neural network architecture [19]–[21]. As a result, these algorithms only work with certain object detection models. In this work, we propose an active learning algorithm that measures data localization informativeness using the temporal information from unlabelled video data, that works with any object detection model.

With separate informativeness scores for the localization and classification component, the most common aggregation methods are sum, average and maximum [18]. In this work we proposed an aggregation method to measure the inconsistency between the localization and classification informativeness using a dynamically changing weight factor and shows that such inconsistency could be a better informativeness measurement than the maximum aggregation in some cases. Since the consecutive frames are likely to contain similar information, we proposed a weight curve to preferably query frames that are far away from annotated frames. By doing this, the localization measurements can also be better estimated in future active learning iterations.

In summary, our contributions are the following:
- We represent two metrics to measure the localization informativeness of video frames, which is compatible with object detection models with regular output format.
- We propose an aggregation function with a dynamically changing balance parameter to aggregate data classification and localization informativeness based on their inconsistency, which is proven to be effective compared to a baseline aggregation approach.
- We demonstrate that the proposed weight curve can efficiently avoid querying adjacent frames from video data in an active learning setting.

## 2 Related Works

**Object detection models** have seen increasing accuracy in recent years. [22] proposed a two-step object detection model: Region Convolutional Network (R-CNN). It first estimates a set of region proposals from the input image (localization), then detects and classifies objects (classification). Fast R-CNN [23] proposed the Region of Interest (ROI) Pooling algorithm that only crops the feature map of the input image, so that it is faster and less computational power consuming compared to R-CNN models. Faster R-CNN [11] introduced the concept of anchors to crop the image more efficiently and designed a region proposal network (RPN) to further speed up the object detector at a higher accuracy. You Only Look Once (YOLO) group of object detection models [10] were first introduced in 2016 and has been improved multiple times since then. They are in a similar architecture as SSD and adapted the anchor concept from Faster R-CNN. The latest YOLO object detector [25] has surpassed R-CNN based models in accuracy while maintains the speed advantage, it has been used for many real-time object detection tasks.

**Active learning on object detection** has attracted more research interest recently. [21], [26] calculates pixel scores and use them for selecting informative samples. [27] works similarly but approximates the uncertainty via MC-dropout. [19] proposes two different measurements: localization tightness which is the overlapping ratio between the region proposal and the final prediction; and localization stability which is based on the variation of predicted object locations when input images are corrupted by noise. [17] proposed black-box and white-box methods. Whereas the black-box methods do not depend on the underlying network architecture, white-box methods are defined based on the network architecture. [28] outperforms the other single model-based methods. During the training, the method learns to predict the target loss for each sample. During the active learning stage, it chooses to label the samples with the highest predicted loss.

**Video object detection models through post processing** utilizes an image object detection model to detect for each frame from the video, then the detections are finetuned using the temporal information from the video data. [29] performed single-frame object detection and object movements tracking across frames in a multi-task fashion. Then it links the detections across frames to object tubelets using the predicted movements, and re-weights detection scores in tubelets. [30] proposed Seq-NMS to form high score linkages using bounding box IoU across frames and then rescore the boxes associated with each linkage to the average or maximum scores of the linkage. The fundamental difference between our proposed active learning algorithm and above models is that our algorithm enables any image object detection model to learn from video data to achieve higher accuracy on any other single image input with similar feature distribution. Moreover, while applying our actively trained object detection model to video data, all the post-processing approaches are compatible.

## 3 Proposed Method

As shown in Figure 2, proposed active learning algorithm uses three score functions to measure data informativeness, the queried data are selected based on their aggregated and weighted scores. To initialize our active learning algorithm for training object detection model using video data, a set of intermediate frames are annotated as the initial training set. These frames will be annotated and used to train the object detection model as the first active learning iteration. For example, using a video with 500 frames; frame 1, frame 50, ..., frame 500 can be selected as the initial training set. The reason that these intermediate frames are selected as the initial training set is that, without any information from the object detection output, the further the distance between the frames the more likely they contain different information.

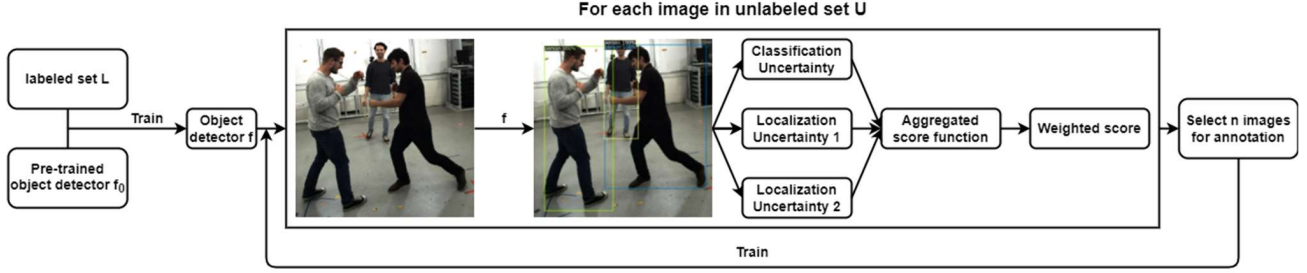

Figure 2. The framework of the proposed algorithm

For the future iterations, the instance bounding boxes and their categories from unannotated frames will be detected using the object detection model trained in the previous iteration. Based on detection outputs, a set of informative frames are queried using our proposed query strategies. In section 3.1, we introduce the score functions that we proposed to measure the data informativeness for both localization and classification component in an object detection model. To combine these scores, we proposed two aggregation functions in section 3.2 to compute the final informativeness score for querying unannotated frames. Queried frames will then be annotated to train the object detection model for next active learning iteration. The proposed algorithm works for any multi-class object detection model. For each object class, data informativeness is measured for each unannotated frame. The class informativeness of the most informative class is used as the informativeness for each frame as suggested by [19]. For simplicity, the following sections covers the definition and formulations using only one class.

## 3.1 Informativeness Measurement – Score Functions

While applying an object detection model $f_\theta$ that detects objects in $k$ classes $\{C_1, ..., C_k\}$ for an $w*h$ image input $d$, the object detection output is in the following form:

$$f_\theta(d) = \{b_j^i, p_j^i\}; \text{ for } j = 1, ..., n_d^k, i = 1, ... k$$
$$\text{where } b_j^i = \{x_j^i, y_j^i, w_j^i, h_j^i\}; \ x_j^i \in [0,w], y_j^i \in [0,h]$$

Where $n_d^k$ is the total number of detected instances that belong to class $k$ in image $d$. $b_j^i$ is the set of detected bounding boxes and each bounding boxes is represented by four coordinates (box $w$, box $h$ and box centre coordinates). Each bounding box is assigned with a class distribution vector $p_j^i$ that represents the estimated probabilities of the bounding box belonging to class 1 to class $k$. Following above notations, while taking an $m$-frame video as input, the object detection output of detected instances in one class is:

$$\{b_n^i, p_n^i\}, \text{ where } i = 1, ..., m; n = 1, ..., n_i \quad (1)$$

where there are $n_i$ instances detected in frame $i$. Each $b_n^i$ represents the $n$th bounding box in frame $i$. Each $p_n^i$ is a class distribution vector shows the probabilities of the bounding box $b_n^i$ belonging to class 1 to class $k$. One single-shot segment of the video is noted as a video piece. In the proposed algorithm, each active learning loop uses one video piece. Using these object detection outputs from video input, three score functions are proposed in the following sections.

### 3.1.1 Classification Uncertainty

To measure the informativeness in a multi-class object detection setting, we adept the uncertainty sampling paradigm and use entropy to measure the data informativeness score for classification. The informativeness of a data example $x$ regarding the current model $P_\theta$ can be measured by its entropy [15]:

$$I_x = -\sum_i P_\theta(y_i|x) \log P_\theta(y_i|x) \quad (2)$$

Where $y_i$ covers all the possible annotations. For object detection models, each data example contains multiple object instances and the entropy for each instance detection can be calculated by Equation (2). There are two approaches to aggregate the entropy of all the detected instances: 1) the average aggregation is more prone to the detection number variance meaning that $n_i$ is more likely to influence the aggregated entropy; 2) the maximum aggregation is more prone to the outlier meaning that the aggregated entropy is dependent on the outliers [18]. In practice, more results have shown that the maximum aggregation performs better in active learning algorithms [18], [19]. In this work, the maximum aggregation will be used for instance informativeness score aggregation. Following the notations in Equation (1), the classification uncertainty score function $C_i$ is shown in Equation (3):

$$C_i = argmax_n - \sum_k p_n^i(k) * log p_n^i(k) \quad (3)$$

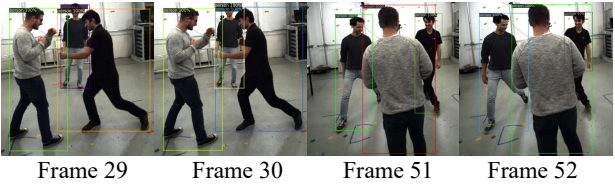

Frame 29    Frame 30    Frame 51    Frame 52

Figure 3: These four images are all frames from a video footage, the left two frames are two consecutive frames that indicates a good localization performance from the model, while the right two consecutive frames show a sign of localization uncertainty.

### 3.1.2 Localization Uncertainty Based on Discontinuity in Detected Instances

Different from the classification uncertainty, the localization uncertainty of a data example cannot be represented with entropy. Because it is impossible to compute the localization probability distribution without heavily modifying the object detector architecture like [20]. Therefore, we use the temporal discontinuity from video data detection output to estimate the localization uncertainty.

For one video piece, the number of instances of one class detected should be a set of continuous integers if the object detection uncertainty in the localization component is 0. Therefore, whenever there is a discontinuity in the number of instances detected over the frames, there is supposed to be an uncertain localization. The estimated instance curve $\tilde{n}_i$ was proposed to detect such discontinuity.

To define $\tilde{n}_i$, the initial training set that contains the intermediate frames are used as the initial guiding set. Based on the number of instances from the annotation of guiding frames, a curve of estimated number of instances in each frame is fitted, which is noted as $\tilde{n}_i$. In the proposed algorithm, the estimated instance curve is fitted by linearly connecting the nodes of each two adjacent guiding frames (other interpolation method are also applicable). By comparing the number of detected instances $n_i$ from each frame to the estimated instance curve $\tilde{n}_i$, the localization uncertainty for frame $i$ can be estimated as follows:

$$\Delta n_i = max\{1, \left|\frac{n_i - \tilde{n}_i}{\tilde{n}_i}\right|\} \quad (4)$$

Note that the percentage difference of $n_i$ compared to $\tilde{n}_i$ could exceed 1 in some situations. Therefore, $\Delta n_i$ is capped at 1 so that it can be treated as a probability value in the aggregation step. After each active learning query step, a set of new frames will be queried based on $\Delta n_i$ after aggregated with other score functions. The selected frames will be annotated and added to the guiding set, so that the fitted $\tilde{n}_i$ curve will become closer to the ground truth.

### 3.1.3 Localization Uncertainty Based on Bounding Box

Similar to $n_i$, the location and size of each detected bounding box should not change drastically from frame to frame for each video piece as shown in Figure 3. Assuming the width of each frame is $w$ and the height is $h$. The detected bounding boxes $b_n^i$ for each frame could be represented by a $w*h$ matrix $H$, named as the localization matrix, the value of each entry is set to 1 indicating at least one detected bounding box is covering the corresponding pixel. We use intersection over union (IoU) to measure the different between the localization matrices of two adjacent frames. Given two matrices with binary entry values, their IoU is defined as:

$$IoU(H_1, H_2) = \frac{H_1 * H_2}{H_1 + H_2}$$

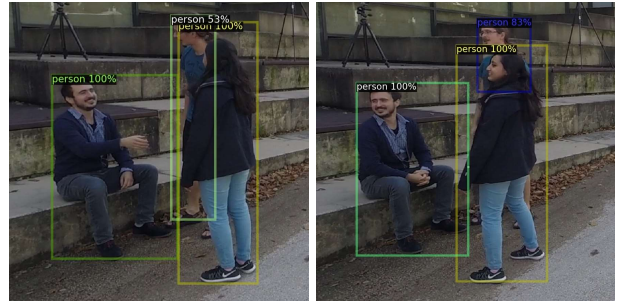

Figure 4: Two image examples of classification uncertainty and localization uncertainty inconsistancy for an object detection model. The detection in the middle of the left image is localized close to the ground truth, while the model is uncertain about whether to categorize it as a person. One the right image, the same person was localized poorly; but based on the captured features from the bounding box, the detector is almost certain that a person was detected.

Like Equation (4), by comparing the localization matrix $H_i$ of frame $i$ to the localization matrices of the adjacent frames, a score can be computed to measure the changes in the size and location of the detected bounding boxes over frames.

$$\Delta H_i = [IoU(H_i, H_{i-1}) + IoU(H_i, H_{i+1})]/2 \quad (5)$$

### 3.2 Score Function Aggregation

While having multiple informativeness scores in an active learning algorithm, there are multiple ways to aggregate these scores. Most used aggregation methods are 1) multiplying the scores, 2) adding all the scores and 3) using the highest score. Even though the value computed from each score function is between 0 and 1 that measures the uncertainty of the detection, the approaches that each of them are computed are drastically different. The classification uncertainty score is computed from the entropy of each instance detection, while the localization scores are computed from the temporal detection discontinuity in a video piece. Taking both localization and classification uncertainty into consideration, [19] suggested that the high inconsistency between localization and classification uncertainty indicates a highly informative example, as shown in Figure 4. Following this concept, an aggregation functions $S_i^1$ is proposed to aggregate $C_i$, $\Delta n_i$ and $\Delta H_i$:

$$S_i^1 = |max\{\Delta H_i, \Delta n_i\} - \mu * C_i| \quad (6)$$

**Algorithm 1**

**Require**: Unlabelled $m$-frame video $U$; Initial object detection model $f_0$; Score functions $C$, $\Delta n$ and $\Delta H$; Aggregation function $S$; Weight curve $w$

**Initialize**:
Query: $G_0 = \{U_0, U_{10}, U_{20}, ...\}$, Construct initial guiding set
Annotation: $G_0^{anno} = (G_0, y_0)$, Label initial guiding set with $y_0$
Train:  $f \leftarrow f_0$ trained with $G_0^{anno}$
    $U \leftarrow U \backslash G_0$, Update unlabelled set
    $L \leftarrow G_0^{anno}$, Construct labelled set

**For** $k$ Iterations:
  **For** $i$ from 1 to $m$: compute detection output $D_i$
    **If** frame $i$ in $L$: $D_i = y_i$
    Fit $\tilde{n}_i$ curve, Update $w_i$
    **If** frame $i$ in $U$: $D_i = f(U_i) = \{b_n^i,\ p_n^i\}$
    $n_i$ = No. of detections in frame $i$
    $H_i$ =  localization matrix for frame $i$
  **While** frame $i$ in $U$:
    Calculate $C_i$, $\Delta n_i$, $\Delta H_i$ and $S_i$
    $\hat{S}_i = S_i \cdot w_i$
  Query: $G_k = \{U_i\}$ s.t. $\hat{S}_i$ is in the 10 highest scores
  Annotation: $G_k^{anno} = (G_k, y_k)$
  Train:  $f \leftarrow f$ trained with $G_k^{anno}$
    $U \leftarrow U \backslash G_k$
    $L \leftarrow L \cup G_k^{anno}$

Since both $\Delta H_i$ and $\Delta n_i$ are computed similarly, $\Delta n_i$ can be considered as a discrete version of the localization uncertainty measurement while $\Delta H_i$ can be considered as continuous. Therefore, the higher value from $\Delta H_i$ and $\Delta n_i$ can be used to represent the localization uncertainty. µ is a balancing parameter that controls how the frames are queried. Large µ means that the frames with high classification uncertainty score are more likely to be queried, whereas small µ means that the frames with high localization uncertainty score are more likely to be queried. For each iteration, µ is calculated as the mean localization uncertainty of unannotated frames divided by the mean classification uncertainty of unannotated frames. To validate the effectiveness of consistency aggregation in Equation (6), a summation aggregation method is also used for comparison:

$$S_i^2 = max\{\Delta H_i, \Delta n_i\} + C_i \qquad (7)$$

For a video piece, adjacent frames will be likely to contain similar information. To avoid repeatedly selecting adjacent frames, a weight curve is proposed to control the likelihood of selecting frames. The weight curve is designed as follows: the weight is set to 0 at each guiding frame and it increases to 1 at the middle point of two closest guiding frames, the weight curve is updated every time after a set of new frames are annotated. To calculate the weighted score, multiply the aggregated score to its corresponding weight. Another benefit of using the weight curve is that the estimated bounding box curve $\tilde{n}_i$ will be fitted more easily, since distanced frames are more likely to be selected for annotation. In the next section, the effectiveness of the weight curve will also be tested.

Putting the aggregation function and the weight curve together, the proposed algorithm is shown in Algorithm 1.

## 4 Experiments

To our knowledge, there is no priorly proposed active learning algorithm for unannotated video data specifically. Following the experiments setup from [19], our proposed active learning algorithm was compared to two baseline query strategies:
- Passive Selection (P): unannotated frames are queried randomly, then added to the labelled set.
- Classification only (C): only use the classification uncertainty as the informativeness measurement.

For all query strategies, ten intermediate frames are annotated as the initial guiding set. This also applies to the passive selection query strategy to ensure that the models are initialized identically. For each active learning query step, ten frames are queried based on the query strategy. Unannotated frames are queried in batches and trained with newly queried frames solely might cause model performance degeneration on previously learned examples. Therefore, the incremental learning scheme was used as suggested in [18]. After the query step, the object detection model is trained for ten epochs with a mini-batch of size twenty. Each mini-batch is constructed by randomly selecting ten frames from the guiding set together with the ten frames queried in the query step. Based on the experiment results, he learning rate was set to 0.001 for consistency.

Proposed active learning algorithm was applied to two object detection datasets to evaluate its performance. Both datasets contain one or multiple video pieces and their details will be covered in the following sections. For each video piece, 10% of the frames will be randomly selected as the test set. The rest of the frames are treated as unannotated frames and will be annotated under different query strategies until 80% of these frames are annotated.

The proposed algorithm is designed to work with any image object detection model. In this work, the model used for object detection is the Faster R-CNN model with a ResNet-50-FPN backbone. It contains two parts: Region Proposal Network (RPN) part for localization and Fast R-CNN part for classification. The RPN part predicts the bounding boxes for possible objects and the Fast R-CNN part classifies these proposed regions and refine the bounding boxes predicted. The output layer is modified to match the object classes for our dataset. Instead of only finetuning the last few layers, all the layers (except backbone) are trained from randomly initialized weights. In this way, the object detection model can learn the localization and classification information from queried frames more efficiently [15].

The modern standard of evaluating an object detection model is using mean average precision (mAP). The detection for one instance is correct if its ground truth intersection over union (IoU) is above a threshold. The average precision is calculated by taking average of the maximum precision at 11 recall values between 0 and 1. The mAP is computed by setting the IoU threshold to multiple values and taking average of the corresponding AP values. To evaluate the performance of proposed active learning algorithm, model mAP on the test set after each active learning iteration is reported. All the mAP values reported are the average of mAP values from three separate runs.

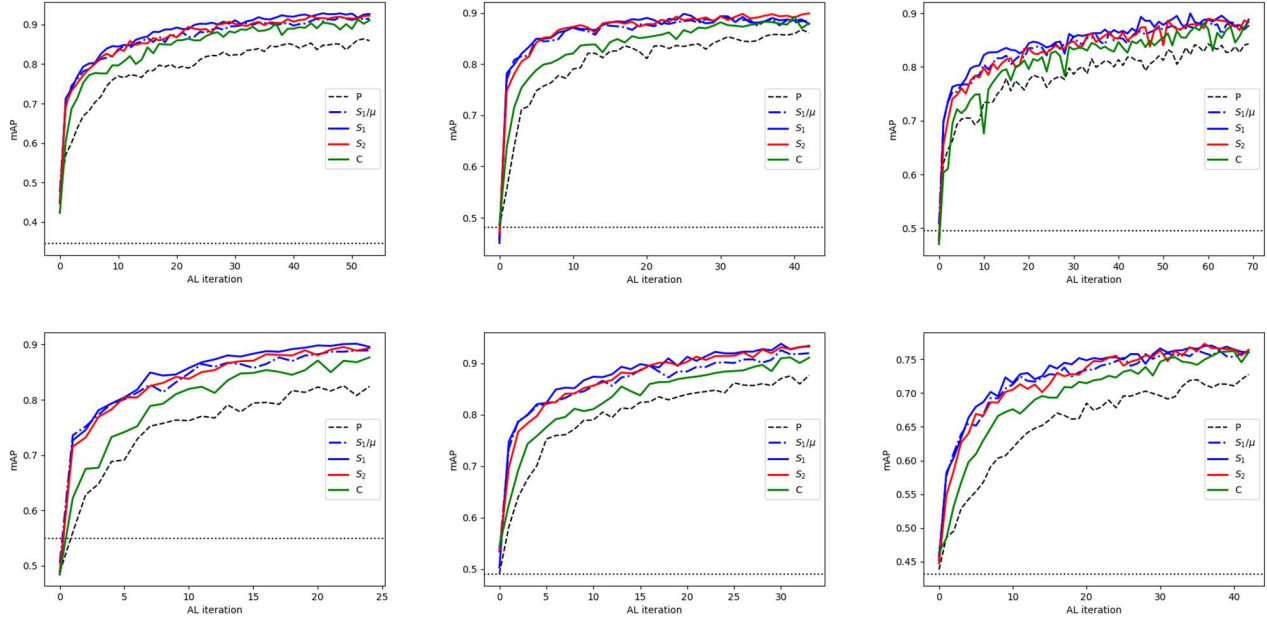

Figure 5: Proposed active learning algorithm performance on six different video pieces. From top left to bottom right: subset 6, 7, 11, 12, 13 and 19. (P) - Passive Selection; (C) – Classification uncertainty only; (S1) – S1 aggregation; (S1/$\mu$) – S1 aggregation with fixed $\mu$; (S2) – S2 aggregation. The horizontal dotted lines are the mAP's using pre-trained Faster R-CNN model.

### 4.1 Multi-person Pose Estimation Test Set

The main experiments were taken out on the Multi-person Pose estimation Test Set (MuPoTS) [1]. It contains twenty subsets, where each subset consists of continuous frames of a video piece. Six subsets were selected for our experiments, the details of selected subsets are shown in supplementary. Besides two baseline query strategies, our proposed algorithm was evaluated in three different configurations. All three configurations take both frame localization and classification informativeness into consideration, using different aggregation functions: 1) S1: S1 aggregation function with changing balancing parameter $\mu$; 2) S1/$\mu$: S1 aggregation function with fixed balancing parameter $\mu = 1$ and 3) S2: S2 aggregation function.

As shown in Figure 5, all active learning query strategies including C outperformed passive selection as expected. As the active learning loop goes on for all subsets, the mAP gain per iteration for S1 and S2 became closer to C. In another word, the training speed improvement of applying proposed active learning algorithm mainly happens at the earlier stage. The video pieces in subsets 6 and 11 are the two longest videos in our experiments and the human body occlusion in these two videos is minimal compared to other video pieces. As a result, including localization uncertainty in query strategies provided less performance gain compared to subsets with heavier human body occlusion. Since object occlusion from input image is the main factor that causes localization uncertainty for the object detection models. Compared to classification uncertainty based active learning algorithms, the advantage of proposed algorithm is less significant for low-occlusion videos.

Comparing two aggregation functions, S1 outperforms S2 in most subsets. For more complex video pieces 7, 13 and 19 (e.g., more objects or heavy occlusion), the improvement of using inconsistency aggregation as in S1 is smaller and became marginal for the later iterations. In subset 7, one person is covering another person for the most part and the consistency aggregation did not work well as a result. Comparing the blue curves to the dotted blue curves, the advantage of the dynamically changing balancing parameter in inconsistency aggregation function can be easily observed. With $\mu$ fixed, S1 and S2 performed similarly.

Based on the results in Figure 5, proposed active learning algorithm for object detection model has better ability to query high informativeness examples compared to classification-uncertainty-based algorithms, especially for videos with more objects and higher occlusion. In general, the inconsistency aggregation and its dynamically changing balancing parameter $\mu$ further boost the performance of proposed active learning algorithm. However, it shows less significant improvement in the video pieces where most frames contain heavy occlusions.

The weight curve effectively improves the performance for all active learning query strategies for video frame annotation (experiment results in supplementary). Without the proposed weight curve, all query strategies tend to select the frames in particular video segments for consecutive active learning iterations. As a result, the training speed is much slower for all query strategies. In some cases, outperformed by passive selection. This is due to frames containing similar information being queried repeatedly for annotation. We also observed that the localization informativeness measurement $\Delta n$ in the early active learning iterations are less accurate. This is because the annotated frames are not separated enough to fit $\tilde{n}_i$ curve accurately. In our experiments, different shaped weight curve performs similar as long as the gradients are not overly large.

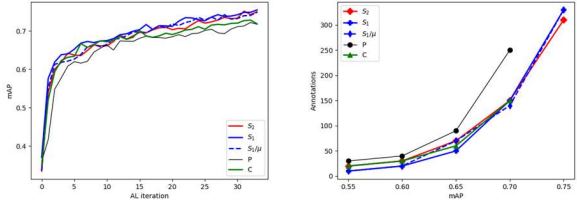

Figure 6: mAP performance evaluation and annotation cost comparison for different query strategies on FootballPD dataset.

### 4.2 Annotation for Football Player Detection Dataset

Object detection is the first step of top-down multi-person 3D pose estimation frameworks [31]. The human object detection accuracy signifyingly affects the accuracy of the localization and estimation of each single-person 3D pose. Using the football broadcast footage as input, one of the major motivations for this work is to use active learning algorithms to train a better player detection model.

To evaluate proposed algorithm in a more realistic setup, we selected a 371-frame video piece from a football broadcast footage and annotated these frames with proposed algorithm. Unlike previous experiments, all queried frames were annotated manually instead of using the ground truth provided in MuPoTS. We want to avoid any machine error during annotation and make sure the sizes of bounding boxes are consistent, since the ground truth bounding boxes of MuPoTS dataset were extracted from a multi-camera system. Compared to the frames in MuPoTS, the instance number in each frame of FootballPD is much higher and the change of instance number from frame to frame is also larger. For each instance, the size of each bounding box is smaller in FootballPD. Besides, there are more complex frames in FootballPD including heavy blurring and occlusion. In general, FootballPD is a harder dataset for object detection, which also makes active learning more challenging

Figure 6 shows the mAP's of the player detection model after 33 active learning iterations and the annotation cost using different query strategies. After annotated and trained the model with 80% of the frames, the player detection mAP reached 75%. The performance improvement of proposed active learning algorithm is consistent with the results on MuPoTS. Both proposed query strategies using aggregation function S1 and S2 reached a higher mAP with lower annotation cost compared to passive selection (P) and classification uncertainty only strategy (C) after 33 iterations. Both P and C query strategies failed to train the model to reach 75% mAP. Moreover, to train the detection model to 70% mAP, our proposed active learning algorithm S1 saved annotation cost by 40%.

## 5 Discussion

We tested the purposed active learning algorithm on two object detection datasets. With MuPoTS, we use the ground-truth annotation to mimic the manually labelling process. The results indicated performance improvement over both passive selection and classification uncertainty only query strategy. Besides MuPoTS, a 371-frame football broadcast footage was annotated with proposed algorithm. For this high-complexity dataset, the experiments showed consistent results as in MuPoTS. All experiment results showed significant reduction in annotation cost while applying proposed algorithm compared to baselines. In active learning settings, the weight curve proposed in this paper was also proven to be effective while querying frames from video data.

The task we want to solve with the proposed algorithm is to efficiently query informative frames from a video for annotation. To evaluate our algorithm on a video piece, the test set is constructed with randomly selected frames from the same video piece. The generalization performance of the actively trained object detection model on other dataset is untested since it is beyond the scope of this work.

All experiments were designed to extract information from single video pieces. However, in a multi-video setup, proposed algorithm could be easily adjusted to query frames simultaneously to construct a multi-video frame batch. This could also be done in a federated learning setting [32], where each video piece is utilized to actively train an object detection model and the model weights will be aggregated iteratively. The benefit of combining federated learning and active learning is that the privacy is preserved since the data annotation and model training are done separately with multiple annotators and only the model weights are centralized for aggregation.

High compatibility is a key advantage of proposed algorithm. To measure data informativeness, all the score functions take the regular object detection output as input. Therefore, proposed algorithm is supposed to work with any object detection model with regular output format.

## 6 Conclusion

To train an object detection model with unannotated video data, labelling each frame is time consuming and insufficient. To select the most informative frames from the video, we propose an active learning algorithm that utilizes the temporal information from video frames to measure the informativeness for each unlabelled frame in the form of three score functions. Proposed algorithm measures both classification and localization informativeness for each frame and is compatible to any object detection model with regular output format. Based on the experiment result, proposed algorithm outperforms baseline active learning algorithms noticeably. It was also demonstrated that, the proposed consistency-based aggregation method together with the weight curve can further reduce the annotation cost. We believe that proposed active learning algorithm is an efficient approach to reduce video annotation cost for object detection, with a great generalization potential to reduce video data annotation cost for more complex computer vision tasks like pose estimation and instance segmentation.